\documentclass[10pt,journal,compsoc]{IEEEtran}
\usepackage{cite}
\usepackage{amsmath,amssymb,amsfonts,dsfont}
\usepackage{algorithmic}
\usepackage[pdftex]{graphicx}
\usepackage{float}
\usepackage{textcomp}
\usepackage{caption}
\usepackage{color,soul}
\usepackage{url}
\usepackage{ulem} 
\usepackage{subfigure}
\usepackage{epstopdf}
\usepackage [latin1]{inputenc}
\usepackage{multirow}
\usepackage{bm}
\usepackage{makecell}
\usepackage[ruled,linesnumbered]{algorithm2e}
\ifCLASSINFOpdf
\else
\fi
\begin{document}
%
\title{ClueGraphSum: Let Key Clues Guide the Cross-Lingual Abstractive Summarization}
\author{Shuyu Jiang,Dengbiao Tu,Xingshu Chen,Rui Tang,Wenxian Wang,Haizhou Wang

\IEEEcompsocitemizethanks{\IEEEcompsocthanksitem Shuyu Jiang, TangRui and Haizhou Wang are with School of Cyber Science and Engineering, Sichuan University, Chengdu 610065, China.
\IEEEcompsocthanksitem Dengbiao tu is National Computer Network Emergency Response Technical Team / Coordination Center of China.
\IEEEcompsocthanksitem Xingshu Chen is with School of Cyber Science and Engineering, Sichuan University, Chengdu 610065, China; Cyber Science Research Institute, Sichuan University, Chengdu 610065, China.
\IEEEcompsocthanksitem Wenxian Wang is with Cyber Science Research Institute, Sichuan University, Chengdu 610065, China.}
\thanks{This work was supported by the Joint Research Fund of China Ministry of Education and China Mobile Company (Grant No. CM20200409) and
the National Natural Science Foundation of China (Grant Nos. 61802270, 61802271)}
\thanks{Corresponding author: Xingshu Chen (chenxsh@scu.edu.cn) and Haizhou Wang (whzh.nc@scu.edu.cn)}
}

\markboth{ }%
{Shell \MakeLowercase{\textit{}}: }

\IEEEtitleabstractindextext{%
\begin{abstract}
Cross-Lingual Summarization (CLS) is the task to generate a summary in one language for an article in a different language.
Previous studies on CLS mainly take pipeline methods or train the end-to-end model using the translated parallel data. However, the quality of generated cross-lingual summaries needs more further efforts to improve, and the model performance has never been evaluated on the hand-written CLS dataset.
Therefore, we first propose a clue-guided cross-lingual abstractive summarization method to improve the quality of cross-lingual summaries, and then construct a novel hand-written CLS dataset for evaluation.
Specifically, we extract keywords, named entities, etc. of the input article as key clues for summarization and then design a clue-guided algorithm to transform an article into a graph with less noisy sentences. One Graph encoder is built to learn sentence semantics and article structures and one Clue encoder is built to encode and translate key clues, ensuring the information of important parts are reserved in the generated summary. These two encoders are connected by one decoder to directly learn cross-lingual semantics.
Experimental results show that our method has stronger robustness for longer inputs and substantially improves the performance over the strong baseline, achieving an improvement of 8.55 ROUGE-1 (English-to-Chinese summarization) and 2.13 MoverScore (Chinese-to-English summarization) scores over the existing SOTA.
\end{abstract}

}

\maketitle

\IEEEdisplaynontitleabstractindextext

%
\IEEEpeerreviewmaketitle

\section{Introduction}

\begin{figure*} [!t]
    \centering
    \includegraphics[width=0.6\textwidth]{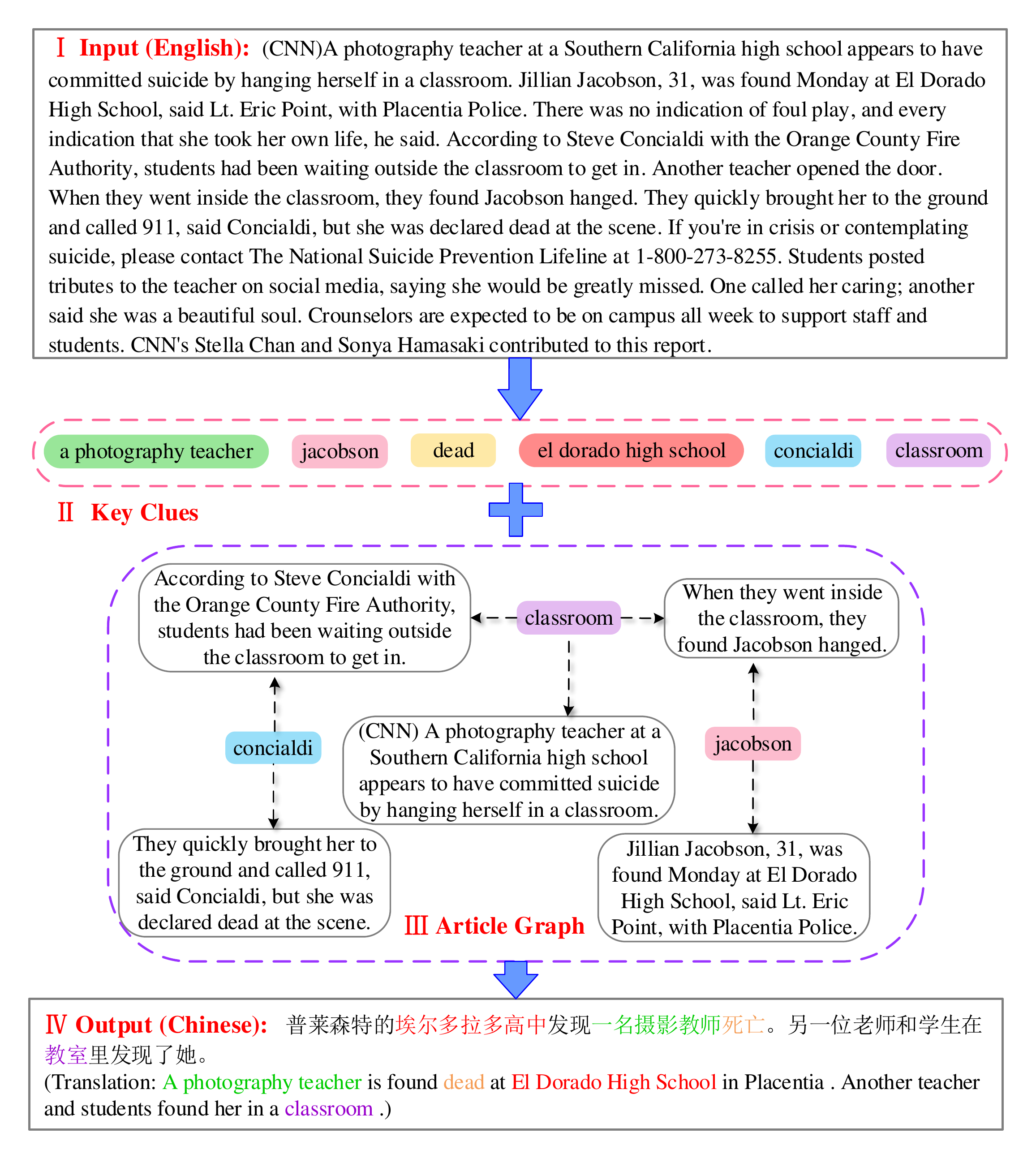}
    \caption{An example of how key clues lead the cross-lingual summarization in our method. Words covered by colored squares represent the key clues for summarization. Sentences containing the same key clue are linked to form an article graph. The colored words in the summary are directly translated from the clues of the same color.}
    \label{pic:exmaple}
\end{figure*}

With the constant acceleration of information globalization, an ever-increasing mass of information in various languages has flooded into people's views ~\cite{aldous2019view, XUAN2017263, NASAR2019102088}. Whether it's scientific research or daily reading, it will always encounter the need to quickly grasp the gists of such massive articles. However, it is hard for people to accomplish this for an article written in an unfamiliar language. At this time, the need for cross-lingual summarization (CLS) has become extremely urgent. Cross-lingual summarization is the task to compress an article in a source language (e.g. English) into a summary in a different target language (e.g. Chinese). It needs to generate a concise text that contains the major points of a given article ~\cite{XIAO2022108483,GUAN2021106973}.

Early research on cross-lingual summarization takes a pipeline-based approach, either summarize-then-translate ~\cite{oravsan2008evaluation, wan2010cross, wan2011using, yao2015phrase,zhang2016abstractive}, or translate-then-summarize ~\cite{leuski2003cross, ouyang2019robust}. Although these methods have adopted various strategies to integrate bilingual features, they still need to train the summary system and translation system separately, suffering from error propagation ~\cite{zhu2019ncls}.

With the widespread use of deep neural networks, many Transformer-based end-to-end methods ~\cite{duan2019zero,zhu2019ncls,zhu2020attend,cao2020jointly, bai2021cross,dou2020deep,takase2020multi} are proposed to directly understand bilingual semantics and avoid the error propagation problem.
Due to lacking large-scale CLS datasets, Duan et al.\cite{duan2019zero} and Dou et al.\cite{dou2020deep} explore training end-to-end models with zero-shot learning. Then Zhu et al.\cite{zhu2019ncls} construct two large-scale cross-lingual summarization datasets and reinforce their Transformer-based neural CLS model (TNCLS) with a multi-task framework. Subsequently, Zhu et al.\cite{zhu2020attend} greatly shorten the training time of multi-task NCLS by adding attending, translating, and summarizing operations into the Transformer-based CLS model instead of using the multi-task framework.

Although these methods has achieved a big improvement, there still is a challenge for models to generate a cross-lingual summary that correctly translates and condenses all major points of a long text, especially when generating the multi-sentence summary.
This is because a long text usually contains much noisy information like organization or author introduction, advertisements, backgrounds, etc. and the attention of models is easily distracted or misled by those.

To handle this problem, we propose a clue-guided cross-lingual abstractive summarizer, \textbf{ClueGraphSum}. The inspiration of \textbf{ClueGraphSum} comes from the pattern how we write cross-lingual summaries.
When writing cross-lingual summaries, we find writers usually tend to select and translate keywords, named entities, etc. as key clues for summarization first, then use related sentences to write target summaries covering these key clues (As shown in Figure \ref{pic:exmaple}).
So we first extract the keywords, named entities, etc. of the input article using TextRank ~\cite{mihalcea2004textrank} as its key clues for summarization; then an article graph with less noisy sentences is constructed based on these clues, capturing the relations and importance of sentences; finally, article graph and key clues are sent to our \textbf{ClueGraphSum} model to jointly decide which clues will be translated and added to CLS summaries. In this process, key clues are used to lock all the important points of articles and article graphs are used to link key clues and supplement them into a complete and fluent summary. As shown in Figure \ref{pic:exmaple}, these key clues and sentences linked by them are informative enough to generate the target summary.

 In conclusion, our contributions are as follows:
\begin{itemize}
\item We propose an article graph construction method that can correctly focus on the most important sentences via the vertex weight we defined and reduce the negative effects of noise by pruning noisy sentences. It transforms an article into a graph constructed of sentences based on key clues.
\item We propose a clue-guided cross-lingual abstractive summarization method, which incorporates article graphs and key clues for summarization into a unified model, namely \textbf{ClueGraphSum}, to co-generate summaries. It utilizes the end-to-end framework to directly map source-to-target language semantics. The experimental results show that our method outperforms the baselines and its performance is comparable with the SOTA. Moreover, our method has stronger robustness for the length of inputs.
\item We constructed a novel hand-written CLS dataset (namely \textbf{CyEn2ZhSum}) to avoid the ``translationese'' ~\cite{graham2020translationese} phenomenon for the evaluation of cross-lingual summarization. It also provides extra reference information for further research.
\end{itemize}
\section{Research objective}
The main research objective of this work is to design an end-to-end model to improve the quality of cross-lingual summaries by enhancing the model's robustness to the article length and unifying the translating and summarizing in one model.
In the previous research about cross-lingual summarization, most of them are pipeline\_based, which suffers from error propagation and costly inference. Thus, we chose to develop an end-to-end model to unify the bilingual semantic mapping and summarizing to avoid this problem.
In the end-to-end research about cross-lingual summarization, the overlength articles are often directly truncated and exist noisy information distracting the model's attention, resulting in the omission of some key points in the target summaries. So, we propose the clue-guided cross-lingual model to alleviate this problem.

In addition, the existing CLS datasets are all obtained through translating the source texts or target summaries with different strategies, which will face the problem of ``translationese''. Therefore, we construct a novel hand-written CLS dataset for evaluation.

\section{Related works}
The overwhelming information globalization forces people to handle vast quantities of information in various languages. Summarizing non-native language documents into native language abstracts provides an effective way to
deal with this situation. Therefore, cross-lingual summarization is continually being researched and optimized in recent years.
\subsection{Pipeline-Based Cross-Lingual Summarization}
Early cross-lingual summarization methods were pipeline\_based ~\cite{leuski2003cross, lim2004multi, oravsan2008evaluation, wan2010cross, wan2011using, yao2015phrase, zhang2016abstractive}, which are translate-then-summarize or summarize-then-translate.

Leuski et al. ~\cite{leuski2003cross} implement the translate-then-summarize method to generate English titles from Hindi documents. Considering the translated text is not completely accurate, Ouyang et al.~\cite{ouyang2019robust} propose to train a robust English summarizer on noisy English documents and clean English summaries. So they can translate the document in low-resource languages into English and then utilize this summarizer to generate English summaries.
Lim et al. ~\cite{lim2004multi} adopt the summarize-then-translate method to obtain a Japanese summary by translating the Korean summary generated by their Korean summarizer. Ora{\v{s}}an et al~\cite{oravsan2008evaluation } first use the maximum Marginal Relevance (MMR) method to summarize Romanian news, and then utilize eTranslator\footnotemark\footnotetext{http://www.etranslator.ro/}, an English-Romanian bidirectional translator, to translate the Romanian summaries into English.
To prevent target summaries from becoming unreadable, Wan et al. ~\cite{wan2010cross} not only simply use the existing translator, they also consider the translation quality of each sentence. They utilize SVM regression to predict the translation quality of input English sentences and then choose informative ones with high translation quality to form the Chinese summaries.

All the above methods only make use of the information from one language side. However, the semantic information from both source and target languages is useful and should be considered ~\cite{wan2011using}.  Wan et al. ~\cite{wan2011using} implement the bilingual information in their graph-based ranking framework to deal with the cross-lingual summarization task. Their experiments' results prove that the information on the target language (Chinese-side) is more helpful.
Inspired by the phrase-based machine translation models ~\cite{koehn2003statistical}, Yao et al.~\cite{yao2015phrase} simultaneously implement sentence extraction and compression based on the bilingual phrases alignment information.
Zhang et al. ~\cite{zhang2016abstractive} resolve source-side documents into predicate-argument structures (PAS) with target-side counterparts, and then produce summarises by fusing bilingual PAS structures.

\subsection{End-to-End Cross-Lingual Summarization}
However, these pipeline-based methods not only result in error propagation ~\cite{peng2021summarising} but also make the inference costly as they need to run a summarization system and a translation system sequentially ~\cite{ladhak2020wikilingua}.

Recently, different end-to-end methods ~\cite{shen2018zero,duan2019zero,zhu2019ncls,zhu2020attend,cao2020jointly, bai2021cross,dou2020deep,ladhak2020wikilingua,maurya2021zmbart,takase2020multi} are proposed to solve these problems.
Due to lacking training corpus, some studies focus on training end-to-end models using zero-shot learning ~\cite{shen2018zero, duan2019zero, dou2020deep} or transfer learning ~\cite{cao2020jointly} and some researchers use multi-task methods ~\cite{takase2020multi,cao2020jointly,zhu2019ncls,bai2021cross} to reinforce cross-lingual summarization with extra parallel data from other related tasks.

Meanwhile, Zhu et al.~\cite{zhu2019ncls} introduce a round-trip translation strategy to construct two large-scale high-quality CLS datasets, respectively En2ZhSum and Zh2EnSum. They then improve the Transformer-based neural CLS model (TNCLS) through a multi-task framework that contains one encoder and two decoders separately performing MT and MS tasks. Cao et al.~\cite{cao2020jointly} use two encoders and two decoders to jointly learn aligning and summarizing.
MCLAS model proposed by Bai et al. ~\cite{bai2021cross} shares one decoder with CLS and MS tasks to establish the connections between discrete phrases in different languages.
Takase et al.~\cite{takase2020multi} further simplify the structure of these multi-task frameworks by attaching a special mark at the beginning of the input sentence to indicate the target task. Their model does not require additional architecture.

Nevertheless, multi-task learning still requires a large amount of extra parallel corpus and its training process is quite time-consuming. To alleviate these issues, Zhu et al. ~\cite{zhu2020attend} integrate TNCLS with an external probabilistic bilingual lexicon through the attending, translating, and summarizing operations. Experimental results show that their method shortens the training time and enhances the model performance at the same time.
To match the training objectives and evaluation objectives of CLS models, Dou et al. ~\cite{dou2020deep} replace the cross-entropy function with the reward mechanism of reinforcement learning (RL). They utilize RL with bilingual semantic similarity as rewards to directly produce a target language summary.

As most methods utilize Transformer as the basic model, their source articles are directly sent or truanted into the model regardless of whether source articles contain noisy information. However, Transformer cannot handle long input all ~\cite{dehghani2018universal} and cannot capture the relation between sentences of source articles and key points of target summaries. To alleviate these problems, we propose a  clue-guiding cross-lingual abstractive summarization method that combines graph networks and the translation pattern into Transformer.

\section{Our Model}
 \begin{figure*} [!ht]
    \centering
    \includegraphics[width=0.75\textwidth]{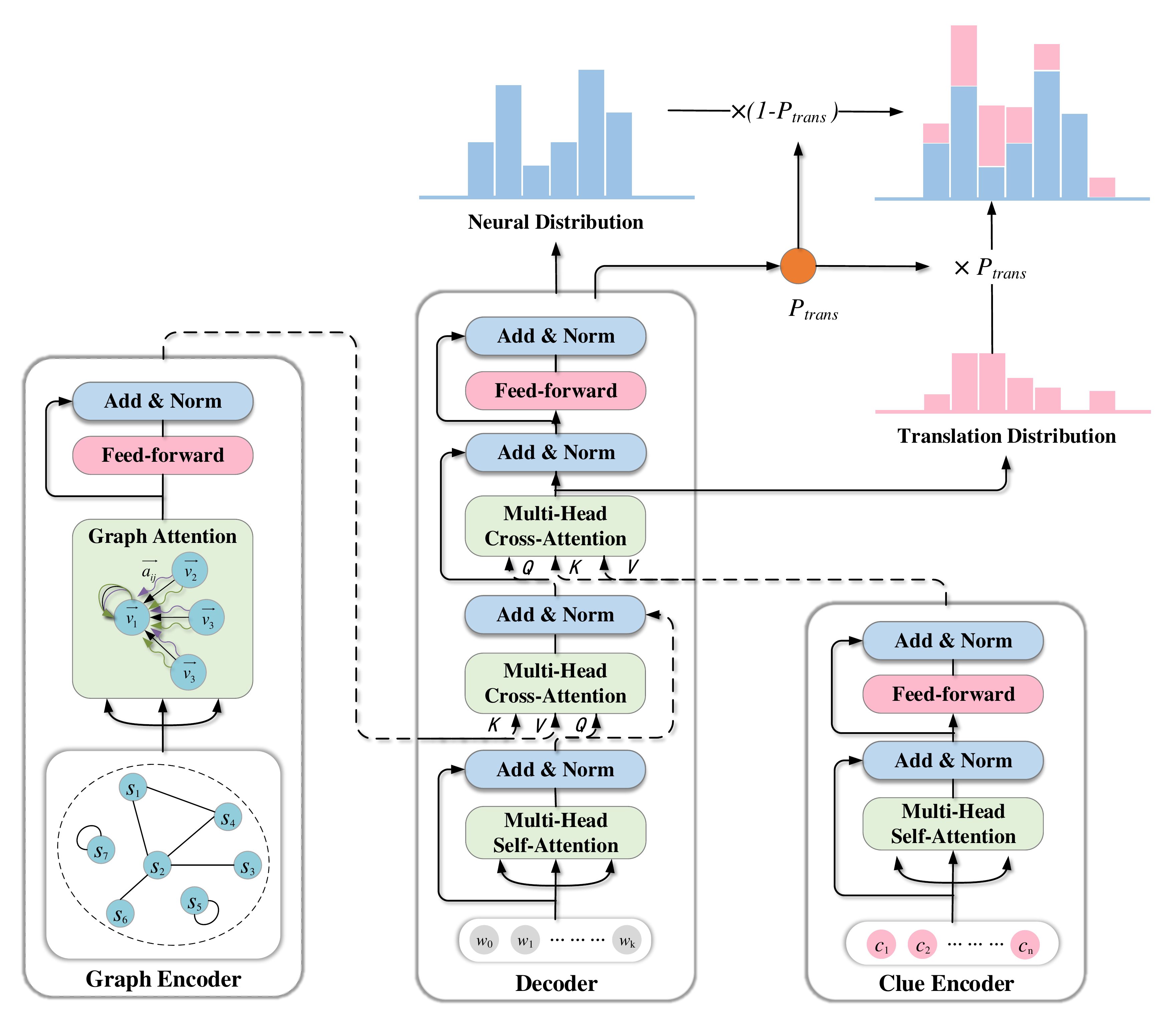}
    \caption{Overview of ClueGraphSum Model. The left part in the picture is Graph encoder, the right part in the picture is Clue encoder and the middle part is the decoder. $s$ represents the embedding of a sentence. $c$ is the key clue and $w$ is the word in target summary. }
    \label{pic:GraphSum}
\end{figure*}

The overall architecture of our model namely \textbf{ClueGraphSum}, is depicted in Figure \ref{pic:GraphSum}.
Here, let $V_{s}$ and $V_{t}$ be the vocabulary of the source language and target language respectively. Given an input article as a sequence of sentences $D=\{ s_{1},s_{2},...,s_{m} \}$ where $s_{i}$ represents a sentence, we first extract keywords and named entities from $D$ as key clues $C=\{ {c_{1},c_{2}...,c_{n}} \}$ using TextRank (~\cite{mihalcea2004textrank}), where $c_{i}$ is a word or phrase ($D,C\subseteq V_{s}$). Then, we construct the article graph $G=(V,E)$ based on $C$ according to Algorithm~\ref{alg:gconstruct} and then embed them into our model as described in Section \ref{sec:graph encode}. The Graph encoder in Section \ref{sec:graph encode} and Clue encoder in Section \ref{sec:Clue encode} relatively maps $G$ and $C$ into hidden vectors. And the decoder generates a summary $S =\{w_{1},w_{2},...,w_{k} \} $ in the target language from them, where $w_{i}$ represent a word and $(w_{i}, S\subseteq V_{t})$.

\textbf{ClueGraphSum} integrates the graph network into Transformer (~\cite{vaswani2017attention}) using the graph attention (GAT) (~\cite{velickovic2018graph}) and adopts the Naive translation strategy ~\cite{zhu2020attend} to decide which words are translated.

\subsection{Encoder}
\subsubsection{Graph Construction}
\label{sec:graph construction}
This section introduces how to construct the article graph based on key clues. Articles extracted from online websites usually contain much noise (~\cite{li2019coherent}), like advertisements, image credits, editorial and author information, etc. Therefore, we first extract keywords, named entities, etc. of the article using TextRank as its key clues to evade these noise.
\begin{algorithm}
\caption{Article Graph Construction}
\label{alg:gconstruct}
\KwIn{The article text $\bm{D}$ and key clues $\bm{C}$}
\KwOut{Article graph $\bm {G}$}
{
     Create empty article graph $G$ \\
    \For{sentence $s$ in $D$ }{
        Add $s$ as vertex $v_{s}$ to $G$ \\
        Add a self-loop $e_{ss}$ to $v_{s}$ \\
        Calculate the weight of self-loop edge $e_{ss}$:
        $\omega_{ss}$ = count(clues in $v_{s}$ ) \\
    }
    \For{$v_{i}$,$v_{j}$ in $G$ }{
        \If{$v_{i}$ and $v_{j}$ contain common clues in $C$ }{
            Add an non-self-loop edge $e_{ij}$ between $v_{i}$ and $v_{j}$ \\
            Calculate the weight of $e_{ij}$:
            $\omega_{ij}$ = count(common clues) \\
        }
    }
    \For{vertex ${v}_{i}$ in the vertexes of $G$ }{
         Calculate the vertex weight:
         $\omega({v}_{i})=\sum_{j\in{neighbor({v}_{i})}}\omega_{ij}+\omega_{ii}$\\
    }
    Remove the vertexes whose weight equals 0 \\
    Return $\bm {G}$\\
}
\end{algorithm}

After getting key clues $C$ of the article, we construct the article graph following Algorithm~\ref{alg:gconstruct} (denoted as \textbf{AGC}). Every sentence is set as a vertex. The vertex $v_{i}$ and $v_{j}$ are connected by edge $e_{ij}$ if they share at least one key clue and the weight of $e_{ij}$ equals the number of key clues they both contain. We also add a self-loop edge for every vertex to enhance attention of self information. The weight of a self-loop edge equals the number of key clues its vertex contains and the weight of every vertex equals the sum of all the connected edges' weights. The importance of an vertex is represented by its weight and the vertex with a weight of 0 is removed to reduce the impact of noisy or useless sentences.

\subsubsection{Graph Encoder}
\label{sec:graph encode}
\textbf{Embedding Vertices:} As stated above, a vertex is associated with a sentence which consists of a word sequence $s=\{ w_{1},w_{2},...,w_{t} \}$. In order to capture the positional information of every word in the sentence, we use Positional Encoding (PE) (~\cite{vaswani2017attention}) to get its positional embedding. The final embedding of $i$-th word is the sum of its original word embedding $\mathbf{w}_{i}$ and its positional embedding. Then we concatenate that as the sentence embedding and feed it into self-attention layer to obtain the embedding $\vec{h}_{s}$ for each vertex $v_{s}$ as follows:
\begin{equation}
\vec{h}_{s}=\text{SelfAttention}\left(\parallel_{i=1}^{t}(\mathbf{w}_{i}+ PE(\mathbf{w}_{i}))\right) .
\end{equation}

\textbf{Embedding Graph:}
After getting the embedding of each vertex, we implement multi-head graph attention to stabilize the learning process and capture the article graph structure as well as sentence information. The graph attention of every head is independently calculated and then concatenated together. In this process, every vertex $v_{i}$ updates its embedding using the attention over itself and 1-hop neighbors represented by the set $\mathcal{N}\left(v_{i}\right)$:
\begin{equation}
\begin{aligned}
\vec{h}_{i}^{\prime}=\|_{f=1}^{F} \sigma\left(\sum_{j \in \mathcal{N}\left(v_{i}\right)} \alpha_{i j}^{f} \mathbf{W}^{f} \vec{h}_{j}\right),
\end{aligned}
\end{equation}

\begin{equation}
\begin{aligned}
\alpha_{i j}=\text{softmax}\left(e_{i j}\right)=\frac{\exp \left(e_{i j}\right)}{\sum_{l \in \mathcal{N}\left(v_{i}\right)}\exp \left(e_{i l}\right)},
\end{aligned}
\end{equation}
\begin{equation}
\begin{aligned}
e_{ij}=\text { LeakyReLU }\left(\boldsymbol{\vec{a}}^{T} \left[\boldsymbol{W}\vec{h}_{i}||\boldsymbol{W}\vec{h}_{j}\right]\right),
\end{aligned}
\end{equation}
where $F$ denotes the number of heads, $\alpha_{i j}$ denotes the attention between vertex $v_{i}$ and $v_{j}$, and $\cdot^{T}$ represents transposition. $\boldsymbol{W}$ and $\vec{a}$ are respectively learnable weight matrix and vector.

In the last layer of graph attention network, averaging is employed instead of concatenation to combine multi-head graph attention outputs:

\begin{equation}
\vec{v}_{i}=\sigma\left(\frac{1}{F} \sum_{f=1}^{F} \sum_{j \in \mathcal{N}\left(v_{i}\right)_{i}} \alpha_{i j}^{f} \mathbf{W}^{f} \vec{h}_{j}^{\prime}\right).
\end{equation}

To avoid the over-smoothing problem, we employ a residual connection (~\cite{he2016deep}) and layer normalization (~\cite{ba2016layer}) around a feed-forward network (FFN) layer to get the final outputs $[\mathbf{z}_{1}^{g},\mathbf{z}_{2}^{g},...,\mathbf{z}_{t}^{g}]$ of the Graph encoder as follows:
\begin{equation}
\mathbf{z}_{i}^{g}= \text{LayerNorm}\left(\vec{v}_{i}+\text{FFN} \left(\vec{v}_{i}\right)\right).
\end{equation}

\subsubsection{Clue Encoder}
\label{sec:Clue encode}
Although the Graph encoder has captured the global article structure and sentence semantics, some importance information are still omitted because a vertex can not represent different word semantics in one sentence. Therefore, we use the encoder of Transformer as \textbf{Clue encoder} to capture the key clues $C$ in sentences. These key clues are concatenated as the input of Clue encoder. After embedding these clues, the hidden vector is fed into a multi-head attention block as query $Q$, key $K$ and value $V$:
\begin{equation}
\begin{aligned}
\text{MultiHead}(Q, K, V) =\left[\text{head}_{1}; \ldots; \text { head }_{\mathrm{h}}\right] W^{O} \\
\text{where head }_{\mathrm{i}} =
\text { Attention }(Q W_{i}^{Q}, K W_{i}^{K}, V W_{i}^{V}),
\end{aligned}
\end{equation}

\begin{equation}
\begin{aligned}
\text{Attention}(Q W_{i}^{Q}, K W_{i}^{K}, V W_{i}^{V})= \\
\operatorname{softmax}\left(\frac{(Q W_{i}^{Q}) (K W_{i}^{K})^{T}}{\sqrt{d_{k}}}\right) (V W_{i}^{V}) \\
\text { where } Q=K=V=\parallel_{i=1}^{n} \left( \mathbf{c}_{i}) \right).
\end{aligned}
\end{equation}
where $W^{O}, W_{i}^{Q}, W_{i}^{K}$, and $W_{i}^{V}$ are learnable matrices, $\sqrt{d_{k}}$ is the dimension of the key and $\mathrm{h}$ is the number of heads.
The results of multi-head attention are passed through the rest layers of Clue encoder as shown in Figure ~\ref{pic:GraphSum} to get the outputs $\mathbf{z}_{i}^{c}$ of Clue encoder.

\subsection{Decoder}
The decoder shares the same architecture as Clue encoder, except that two extra cross-attention layers are added to perform multi-head attention on the outputs of Graph encoder and Clue encoder.
Considering that some keywords in the cross-lingual summary can be directly translated from the key clues, we leverage the Naive strategy proposed by Zhu et al.\cite{zhu2020attend} to acquire the translation probability $P_{\mathrm{T}}$, deciding which words are directly translated. The translating probability $p_{\text {trans} \in[0,1]}$ is computed from the decoder hidden state $H_{dec}$ via a dynamic gate:

\begin{equation}
p_{\text {trans }}=\sigma\left(\mathbf{W}_{2}\left(\mathbf{W}_{1} \mathrm{H}_{\mathrm{dec}}+b_{1}\right)+b_{2}\right),
\end{equation}
where $b_{1}$ and $b_{2}$ are learnable bias vectors, $\mathbf{W}_{1}$ and $\mathbf{W}_{2}$ are learnable matrices, and $\sigma$ is the sigmoid function.

The $p_{\text {trans}}$ determines whether to generate a word $w$ from the neural distribution sampling $P_{\mathrm{N}}(w)$ or directly select a word from the translation candidates of the source words. Thus, the final probability distribution $P(w)$ is calculated from $P_{\mathrm{N}}(w)$, $P_{\mathrm{T}}$ and $p_{\text {trans}}$ as follows:
\begin{equation}
\begin{array}{r}
P(w)=p_{\text {trans }} \sum_{i: w_{i}=w_{\text {src }}} \alpha_{t, i} P_{\mathrm{T}}\left(w_{\mathrm{src}} \Rightarrow w\right) \\
+\left(1-p_{\text {trans }}\right) P_{\mathrm{N}}(w).
\end{array}
\end{equation}
where $P_{\mathrm{T}}\left(w_{\mathrm{src}}\Rightarrow w\right)$ is the translation probability of word $w_{\mathrm{src}}$ to word $w$.

\section{Experimental Setup}
\subsection{Datasets}
We evaluate our method on the benchmark CLS datasets \textbf{En2ZhSum} and \textbf{Zh2EnSum} created by Zhu et al. \cite{zhu2019ncls} and our human-written CLS dataset \textbf{CyEn2ZhSum}.

En2ZhSum is an English-to-Chinese summarization dataset converted from MSMO ~\cite{zhu2018msmo} and CNN/DailyMail ~\cite{hermann2015teaching}. Zh2EnSum is a Chinese-to-English summarization dataset constructing from LCSTS ~\cite{hu2015lcsts} dataset. The summaries of these two datasets are both translated from original monolingual summaries with a round-trip translation strategy.
Since the translated texts are often ``translationese'', we construct a small but precise human-written dataset, namely \textbf{CyEn2ZhSum}, to further evaluate performance of models on human-written CLS.

\begin{table}[h]
  \centering
\caption{The statistics of CyEn2ZhSum. $Ctitle, Etitle, Doc and Sum$ represent the Chinese title, the English title, the source article and the target summary of CyEn2ZhSum dataset. $AvgChars$ and $AvgSents$ respectively represent the average number of characters and sentences. $AvgSentsChar$ refers to the average number of characters in a sentence. $TotalNum$ is the total number of CLS pairs. }
 \begin{tabular}{ccccc}
  \hline
  \textbf{Statistic} &\textbf{Ctitile} & \textbf{Etitle}& \textbf{Doc}&\textbf{Sum}\\
  \hline
         AvgChars&20.27& 11.37&593.60&135.81\\
        AvgSents&1.01  & 1.06 &19.27&3.69\\
        AvgSentsChar&20.17  & 10.84 &35.45&36.89\\
  \hline
      TotalNum & \multicolumn{4}{c}{3600}\\
  \hline
  \end{tabular}
 \label{tab_exp:cy}
\end{table}

The input of CyEn2ZhSum is an English news article about cybersecurity, and the output is a Chinese summary which is first written by a postgraduate, and then reviewed and revised by researchers. In addition, CyEn2ZhSum also contains extra reference information like titles in Chinese and English, publish dates, etc.
The statistics of CyEn2ZhSum dataset are presented in Table ~\ref{tab_exp:cy} and the division of these datasets is presented in Table ~\ref{ex:dataset}.

\begin{table}[h]
 \centering
 \caption{The division details of datasets}
 \begin{tabular}{ccccc}
 \hline
 \textbf{Datasets}&\textbf{Source}&\textbf{Train}&\textbf{Valid}&\textbf{Test}\\
 \hline
 En2ZhSum&News&364,687&3,000&3,000\\
 Zh2EnSum&Microblog&1,693,713&3,000&3,000\\
 CyEn2ZhSum&News&3000&300&300\\
 \hline
 \end{tabular}
  \label{ex:dataset}
\end{table}
An example of CyEn2ZhSum dataset is provided in Appendix~\ref{ap:1}.

\subsection{Baselines}
We compare our method with two relevant pipeline-based methods and several strong end-to-end methods:

\begin{itemize}
\item \textbf{Pipe-TS}: It refers to the translate-then-summarize method. It first translates the source text into the target language by Google Translator\footnotemark\footnotetext{https://translate.google.com/}. Due to the lack of corresponding MS datasets in the target language, we then following Refs ~\cite{zhu2020attend,zhu2019ncls,dou2020deep,xu-etal-2020-mixed} summarize the translated text by LexRank ~\cite{erkan2004lexrank}, a powerful unsupervised summarization method.
\item \textbf{Pipe-ST}: It refers to the summarize-then-translate method. It first trains a monolingual summarization model on monolingual pairs using Transformer then translates the summary into the target language by Google Translator.
\item \textbf{TNCLS}: It refers to the Transformer-based Neural Cross-Lingual Summarization model proposed by Ref \cite{zhu2019ncls}, which is an end to end automatic bilingual abstractive summarization system.
\item \textbf{ATS-NE}: It refers to the ATS model with Naive strategy, which is an end-to-end method proposed by Ref \cite{zhu2020attend}. ATS model integrates the operations of Attending, Translating and Summarizing by adding an additional translation layer to Transformer. In the translating operation, the authors propose 3 strategies, of which we use the Naive strategy.

\item \textbf{ATS-A}: It refers to the end-to-end ATS model ~\cite{zhu2020attend} with Adapt strategy. As the Adapt strategy performs best in its translation strategy, we choose it as our baselines as well.
\end{itemize}

We denote our \textbf{ClueGraphSum} model as \textbf{CGS}.
\begin{table*}[!htbp]
\centering
\caption{Experimental results of each method on the En2ZhSum dataset and Zh2EnSum dataset. The best results of baselines are underlined with wavy lines and those of all models are shown in bold. R1, R2, RL, ME and MS represent ROUGE-1, ROUGE-2, ROUGE-L, METEOR and MoverScore respectively}
 \begin{tabular}{ccccccccccc}
    \hline
    \multirow{2}{*}{\textbf{Method}}
    &\multicolumn{4}{c}{\textbf{En2ZhSum}}
    &\multicolumn{5}{c}{\textbf{Zh2EnSum}}\\
    \cline{2-5}
    \cline{7-11}
    & \textbf{R1} & \textbf{R2} & \textbf{RL} & \textbf{ME}  & &\textbf{R1} &\textbf{R2}& \textbf{RL} & \textbf{ME} &\textbf{MS}\\
    \hline
    Pipe-TS&23.58 & 8.28 & 18.29 & 8.32 & &18.92 & 5.69 & 17.03 & 11.84 &12.18\\
    Pipe-ST&27.23 & 8.91 & 19.50& 11.25  & &31.39 & 14.28 & 29.15 & 16.00&25.66\\
    TNCLS& 34.93& 15.04 & 27.13 & 14.94  & &35.21 &18.59 & 33.01& 17.82&28.74\\
    ATS-NE&36.94 & 17.23 & 28.25& 15.48 & &36.31 & 20.12 & 34.35 & 18.04&29.85\\
    ATS-A& \uwave{38.01} & \uwave{18.37} & \uwave{29.37} & \uwave{16.51} & &\uwave{37.41} & \uwave{21.08} & \uwave{35.60}& \uwave{18.54}& \uwave{30.66}\\
    \hline
    CGS&\textbf{46.56} & \textbf{23.97} & \textbf{32.86}&\textbf{20.72} & &\textbf{40.80} & \textbf{22.05} & \textbf{37.88}& \textbf{20.69} &\textbf{32.79}\\
    \hline
  \end{tabular}
\label{tab_exp:En2ZhSum}
\end{table*}

\subsection{Evaluation Metrics}
Following the previous studies ~\cite{duan2019zero, cao2020jointly, zhu2019ncls, zhu2020attend, ERMAKOVA20191794, see2017get}, we assessed all models with the F1 scores of standard ROUGE metrics ~\cite{lin2004rouge} (ROUGE-1, ROUGE-2 and ROUGE-L) and METEOR metric ~\cite{denkowski2014meteor} on all datasets. And the ROUGE scores of Chinese outputs are calculated on character-level ~\cite{hu-etal-2015-lcsts}.
Besides, we employ MoverScore\footnotemark\footnotetext{https://github.com/AIPHES/emnlp19-moverscore} ~\cite{zhao-etal-2019-moverscore} as additional evaluation metric for English summaries, because it compares the similarity of two samples based on their semantics rather than the word repetition rate.

 \textbf{ROUGE:} ROUGE-1/2/L evaluates the quality of generated summary by calculating the co-occurrence probability of units (unigrams, bigrams and the longest common sequence respectively) in the generated summary and reference summary. The F1 score of ROUGE metrics is defined as follows:
\begin{equation}
R_{R}=\frac{Count(overlappped\_units)}{Length(ReferenceSummary)}
\end{equation}
\begin{equation}
P_{R}=\frac{Count(overlappped\_units)}{Length(GeneratedSummary)}
\end{equation}
\begin{equation}
F_{R}=\frac{2*R_{R} P_{R}}{R_{R}+ P_{R}}
\end{equation}
where $R_{R}$,$P_{R}$ and $F_{R}$ are respectively the Recall, Precision and F1 score of ROUGE. For ROUGE-1/2, the length of summary equals the number of unigram/bigram in summary. Following Hu et al. ~\cite{hu-etal-2015-lcsts}, the ROUGE score of Chinese summaries is calculated by character.

\textbf{METEOR:} METEOR metric expands the clue of ``co-occurrence". Apart from the absolutely identical unit, it also counts words with the same stem and synonyms into the co-occurrence unit, making it more relevant to human judgments. At the same time, METEOR metric takes the word order into the assessment category and establishes a penalty mechanism based on it.
The METEOR score is computed as follows:
\begin{equation}
\text { Meteor }=(1-\text { Penalty }) \cdot F_{M}
\end{equation}
\begin{equation}
F_{\text {M }}=\frac{P_{M} R_{M}}{\alpha P+{M}+(1-\alpha) R_{M}}
\end{equation}
\begin{equation}
\text { Penalty }=\gamma\left(\frac{\text {\# chunks }}{\text { \# unigrams\_matched }}\right)^{\theta}
\end{equation}
where $F_{M}$ is the F1 score of METEOR; $Penalty$ is the penalty factor for word order; $\alpha$, $\gamma$ and $\theta$ parameters are set according to the official settings \footnotemark\footnotetext{http://www.cs.cmu.edu/~alavie/METEOR/}.

$R_{M}$ and $P_{M}$ are the unigram Recall and unigram Precision, which are obtained in the same way as ROUGE. $\#chunks$ represents the number of overlapped chunks, $\#unigrams\_matched$ represents the number of overlapped unigrams in the reference summary.

\textbf{MoverScore:}
MoverScore is a semantic-based evaluation metric for text generation tasks. It encodes the candidate and reference by contextualized word embeddings, and then leverage the Earth Mover's Distance ~\cite{rubner2000earth} to compute the semantic similarity via comparing their embeddings.

\subsection{Experimental Settings}
\label{ex_set}
Following configuration $transformer\_base$ \cite{vaswani2017attention}, the embedding dimensions $d_{model}$ and the inner dimension $d_{ff}$ are respectively set to 512 and 2048, and Clue encoder and decoder both have 6 layers of 8 heads for attention. For Graph encoder, it is set to 1 layer of 3 heads for attention. We use the batch size of 1024 for all datasets. The vocabulary setting follows Ref \cite{zhu2019ncls}.

All the parameters are initialized via Xavier initialization method ~\cite{glorot2010understanding} and the dropout rate is 0.1.
During training, we apply Adam ~\cite{kingad2015methodforstochasticoptimization} with $\beta_{1}=0.9$, $\beta_{2}=0.998$ and $\epsilon = 10^{-9}$ as the optimizer to train parameters.
During the evaluation stage, we decode the sequence using beam search with a width of 4 and length penalty of 0.6.
For CyEn2ZhSum dataset, we carry out the experiments of several end-to-end methods (including TNCLS, ATS-NE, ATS-A, CGS) on it through fine-tuning the model trained from En2ZhSum dataset.

\section{Results and Analysis}
\subsection{Automatic Evaluation}

\textbf{Results on En2ZhSum and Zh2EnSum.}
We have replicated baselines on the En2ZhSum and Zh2EnSum datasets. The experimental results are provided in Table ~\ref{tab_exp:En2ZhSum}. As shown in Table ~\ref{tab_exp:En2ZhSum}, our model CGS can substantially outperform all baselines in both English-to-Chinese and Chinese-to-English summarization. It achieves an improvement of 3.39+ ROUGE-1 and 2.51+ METEOR scores compared with baselines.
Compared with ATS-NE model which utilizes the same translation strategy with CGS, CGS obviously performs better in all metrics.
Compared with ATS-A model which is the best model in baselines, CGS achieves an improvement of 5.60 ROUGE-2 score on the En2ZhSum dataset and 2.13 MoverScore score on the Zh2EnSum dataset. This reveals our method is superior to baselines not only in word repetition, but also in semantics.

It is clear that the performance of end-to-end methods exceeds that of pipeline methods in both directions although the pipeline methods utilize the most advanced machine translation system. This is because end-to-end methods could directly establish the mapping relation between two languages, avoiding the error propagation. Further comparing the performance of two pipeline methods, we find the summarize-then-translate method outperforms the translate-then-summarize method in both cases, which is consistent with the conclusions of research (~\cite{wan2010cross,cao2020jointly}).
Especially, the late-translation method far exceeds the early-translation one on the Zh2EnSum dataset with 12.37 ROUGE-1 and 13.48 MoverScore scores. We speculate this is because the Zh2EnSum corpus has higher requirements for summarization ability (as its word repetition rate of summaries and references is much lower than En2ZhSum dataset), while the early-translation method already causes the semantic deviation at the first stage.

\begin{table}[!htbp]
    \centering
    \caption{ROUGE F1 and METEOR scores of end-to-end methods (TNCLS, ATS-NE, ATS-A and CGS) on the CyEn2ZhSum dataset.}
    \begin{tabular}{cccccc}
    \hline
    &\textbf{Method} & \textbf{R1} & \textbf{R2}& \textbf{RL} & \textbf{ME}\\
    \hline
          &TNCLS&37.49 & 18.69 & 32.65& 16.12\\
          &ATS-NE&39.01 & 20.86 & 34.91 & 17.16\\
          &ATS-A& 39.18& 21.43 & \textbf{36.72} & 17.49\\  
          &CGS &\textbf{42.91}&\textbf{22.67}& 35.36&\textbf{21.33}\\
    \hline
    \end{tabular}
    \label{tab_exp:CyEn2ZhSum}
\end{table}
\textbf{Results on CyEn2ZhSum.}

\begin{table*}[!htbp]
\centering
\caption{Ablation study for the different component of CGS model.  ``CGS-C'' refers to the variant of CGS model that only contains Clue encoder and decoder. ``CGS-G'' refers to the variant of CGS model that only contains Graph encoder and decoder.  }
 \begin{tabular}{m{1cm}<{\centering}m{0.6cm}<{\centering}m{0.6cm}<{\centering}m{0.6cm}<{\centering}m{0.6cm}<{\centering}cm{0.6cm}<{\centering}m{0.6cm}<{\centering}m{0.6cm}<{\centering}m{0.6cm}<{\centering}m{0.6cm}<{\centering}cm{0.6cm}<{\centering}m{0.6cm}<{\centering}m{0.6cm}<{\centering}m{0.6cm}<{\centering}}
    \hline
    \multirow{2}{*}{\textbf{Method}}
    &\multicolumn{4}{c}{\textbf{En2ZhSum}}
     &\multicolumn{1}{c}{}
    &\multicolumn{5}{c}{\textbf{Zh2Ensum}}
     &\multicolumn{1}{c}{}
    &\multicolumn{4}{c}{\textbf{CyEn2ZhSum}}\\
    \cline{2-5}
    \cline{7-11}
    \cline{13-16}
    & \textbf{R1} & \textbf{R2} & \textbf{RL} & \textbf{ME}  & &\textbf{R1} &\textbf{R2}& \textbf{RL} & \textbf{ME} &\textbf{MS} & & \textbf{R1} & \textbf{R2} & \textbf{RL} & \textbf{ME} \\
    \hline
    CGS-G&21.10 & 3.66 & 14.71& 8.22  & &18.30 & 2.93 & 15.76 & 7.87 &15.45 & &31.33 &14.43 &27.77 &15.38\\
    CGS-C&44.48 & 22.42 & 30.44& 18.96 & &38.22  & 19.20 & 36.25 & 19.18 &30.66 &  &37.13&19.20&32.09&15.96\\
   CGS&\textbf{46.56} & \textbf{23.97} & \textbf{32.86}&\textbf{20.72} & &\textbf{40.80} & \textbf{22.05} & \textbf{37.88}& \textbf{20.69} &\textbf{32.79} & &\textbf{42.91}&\textbf{22.67}& \textbf{35.36}&\textbf{21.33}\\
    \hline
  \end{tabular}
\label{tab_exp:ablation}
\end{table*}

\begin{table*}[!thbp]
  \centering
\caption{ The human evaluation results of summaries from 3 aspects: informativeness (IF), conciseness (CC) and fluency (FL). The values of Fleiss' Kappa of these three aspects are respectively 0.34, 0.44 and 0.47.}
 \setlength{\tabcolsep}{1.5mm}{
  \begin{tabular}{cccccccccccccccc}
  \hline
  \multirow{2}{*}{\textbf{Method}}
  &\multicolumn{3}{c}{\textbf{En2ZhSum}}
  &\multicolumn{1}{c}{}
  &\multicolumn{3}{c}{\textbf{Zh2Ensum}}
  &\multicolumn{1}{c}{}
  &\multicolumn{3}{c}{\textbf{CyEn2ZhSum}}\\
  \cline{2-4}
  \cline{6-8}
  \cline{10-12}
   & \textbf{IF} & \textbf{CC}& \textbf{FL} & & \textbf{IF}& \textbf{CC}& \textbf{FL} & & \textbf{IF} & \textbf{CC}& \textbf{FL}\\
  \hline
  TNCLS&6.36 & 6.19 & 6.75 & &6.34 & 8.02 & 8.12 & &6.40 &6.82 &6.96 \\
  ATS-NE&6.72 & 6.45 & 6.76 & &6.98 & 8.11 & 8.25 & &7.20&7.43&7.57\\
  ATS-A&6.80 & 6.90 & 7.06 & &7.24 & 8.20 & 8.45 & &7.75&7.71&7.71\\
  CGS&\textbf{7.35} & \textbf{6.99} & \textbf{7.20} & & \textbf{7.51} & \textbf{8.25} & \textbf{8.55} & & \textbf{7.91} &\textbf{7.85}&\textbf{8.50} \\
  \hline
  \end{tabular}}
  \label{tab_exp:human}
 \end{table*}
We implement end-to-end baselines and our method on the human-written CyEn2ZhSum dataset. The experimental results are shown in Table ~\ref{tab_exp:CyEn2ZhSum}.
It is obvious that CGS can achieve better performance in all metrics except for ROUGE-L. As the CyEn2ZhSum dataset belongs to the cybersecurity domain, it contains lots of proper nouns in the field of cybersecurity. The translation of these proper nouns are much different from daily words. Thus, we speculate that the higher ROUGE-L score of ATS-A is attributed to its Adapt strategy (~\cite{zhu2020attend}) which can appropriately adjust the translation probability of the source-target pairs to the new domain. However, the Adapt strategy requires more parameters although it is better than the Native strategy we adopted.

The higher METEOR and MoverScore scores of CGS indicate the overall semantics of summaries generated by CGS is closer to reference summaries although more the longest common sequence appear in the summaries of ATS-A. This is most likely because some reference words in the summary generated by CGS are represented by synonyms or the same morpheme with different forms.

In conclusion, our CGS model markedly outperforms baselines in terms of automatic evaluation.
\subsection{Ablation study}

In this section, we carry out an ablation study to analysis the importance of each key component of our method. Graph encoder and Clue encoder are respectively removed in the ablated version of CGS.
Table \ref{tab_exp:ablation} presents the experimental results of this ablation study on En2ZhSum, Zh2EnSum and CyEn2ZhSum datasets. Results reveal that both ablation operations will cause a decrease in the model performance.
We find Clue encoder contributes more to our model than Graph encoder as CGS-C performs much better than CGS-G. This is because Graph encoder only captures sentence semantics while Clue encoder learns multiple key clues, capturing richer and more core semantic information.

\subsection{Human Judgement}
ROUGE and METEOR can evaluate the quality of summaries from the perspective of word coverage, but it is difficult to evaluate the conciseness, informativeness and fluency of the generated summaries. So we randomly selected 60 samples form the test sets of all the datasets for manual evaluation.

 \begin{table}[!htbp]
    \centering
 \caption{The overall evaluation of summaries by humans. Its Fleiss' Kappa score is 0.67.}
 \setlength{\tabcolsep}{0.45mm}{
    \begin{tabular}{ccccc}
    \hline
    &\textbf{Method} & \textbf{En2ZhSum} & \textbf{Zh2Ensum}& \textbf{CyEn2ZhSum}\\
    \hline
          &TNCLS&-0.15 & -0.35 &-0.28 \\
          & ATS-NE&-0.14 & -0.05 &-0.06\\
          & ATS-A&-0.03& 0.01 &-0.01\\
          &CGS&\textbf{0.32}& \textbf{0.39} &\textbf{0.35}\\

    \hline
    \end{tabular}}
    \label{tab_exp:OE}
\end{table}

Three researchers are firstly asked to evaluate the quality of generated summaries from three independent perspectives: \textbf{conciseness}, \textbf{informativeness}, and \textbf{fluency}. Each perspective is scored from 1 (worst) to 10 (best). The average scores are given in Table ~\ref{tab_exp:human}.
They then compare the overall proximity of generated summaries to reference summaries using Best-Worst Scaling method (~\cite{kiritchenko2017Best}). The results are shown in Table ~\ref{tab_exp:OE} and its values range from 1 to -1. A higher value indicates a model with better performance.

In addition, we leverage Fleiss' Kappa (~\cite{fleiss1971measuring}) to evaluate the scoring consistency between annotators.

As shown in the Table ~\ref{tab_exp:human}, our method obviously performs better than baselines in informativeness, conciseness and fluency on all datasets. What's more, the results in Table ~\ref{tab_exp:OE} indicates the summaries produced by our CGS are the closest to the reference summaries, which is the same with automatic evaluation results.

\subsection{Can the AGC Algorithm focus on the importance sentences? }
To further explore whether our AGC algorithm can really focus on the right important information, we compare it with two other graph-based summary extraction algorithms, respectively TextRank and LexRank.
\begin{table}[!htbp]
  \centering
\caption{The results of different graph-based extraction methods.}
  \begin{tabular}{ccccc}
  \hline
  &\textbf{Method} & \textbf{RL} & \textbf{ME}& \textbf{MoverScore}\\
  \hline
        & TextRank&19.01& 11.70 &13.44\\
        & LexRank& 18.27 & 12.14 &13.34\\
        &AGC&\textbf{23.20}& \textbf{14.83} &\textbf{15.64}\\
  \hline
  \end{tabular}
  \label{tab_exp:aga}
\end{table}

We use these three methods to extract the most important three sentences as summaries, and then evaluate their performance with the ROUGE-L, METEOR and MoverScore metrics.
We conduct this experiment on En2ZhSum dataset due to the length of articles. The results in Table ~\ref{tab_exp:aga} show our AGC algorithm outperforms these two comparative methods. Thus our AGC algorithm can correctly focus on the sentences that are most important to the summary.

\subsection{The Impact of Article Length}
\begin{figure} [!htbp]
  \centering
  \includegraphics[width=0.47\textwidth]{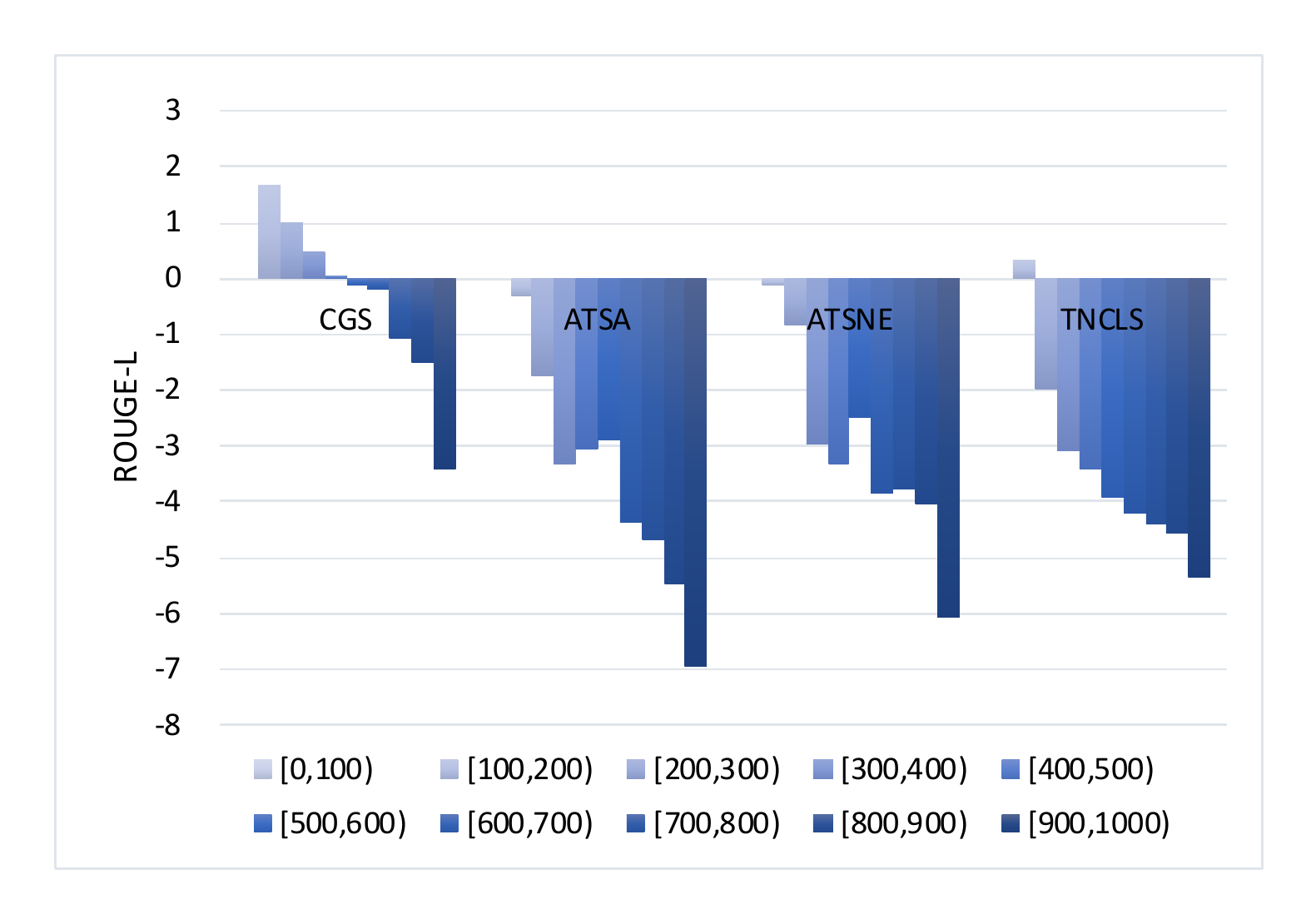}
   \caption{ROUGE-L score difference of summaries generated from articles of different lengths. The legend indicates the range of the number of words in the article. Vertical coordinates represents the difference of Rouge-L scores between each set and the [0,100) set. }
  \label{exp:src_length}
\end{figure}

In order to explore the impact of article length on each model, we divide the articles in the En2ZhSum test set into 10 collections according to the article length. These 10 collections are shown in the legend in Figure ~\ref{exp:src_length}, where $[a, b)$ represents a set of articles whose number of words is bigger than or equal to $a$ and smaller than $b$. We compare the changes in ROUGE-L scores of end-to-end baselines (including CGS, ATSA, ATSNE, TNCLS) on these 10 collections. Based on the results of each model on the $[0,100)$ collection, the score differences between other sets and this set are shown in Figure ~\ref{exp:src_length}.

 Results in Figure ~\ref{exp:src_length} show that the ROUGE-L score of each model decreases with the increase of article length, but our CGS decreases much slower than other models. This indicates that CGS is more robust to the length of input texts. And the overall downward trend most likely results from the increasing redundant information distracting the attention of models.
The reason why our CGS can greatly relieve the impact of the change of article length on the model is that it extracts the important information as key clues to emphatically focus on in the first step and the noisy sentences and important sentences are treated differently at the time of graph encoding. So the attention of CGS model is largely undistracted by useless information.

\subsection{Case Study}
\begin{figure*} [!htbp]
  \centering
  \includegraphics[width=0.85\textwidth]{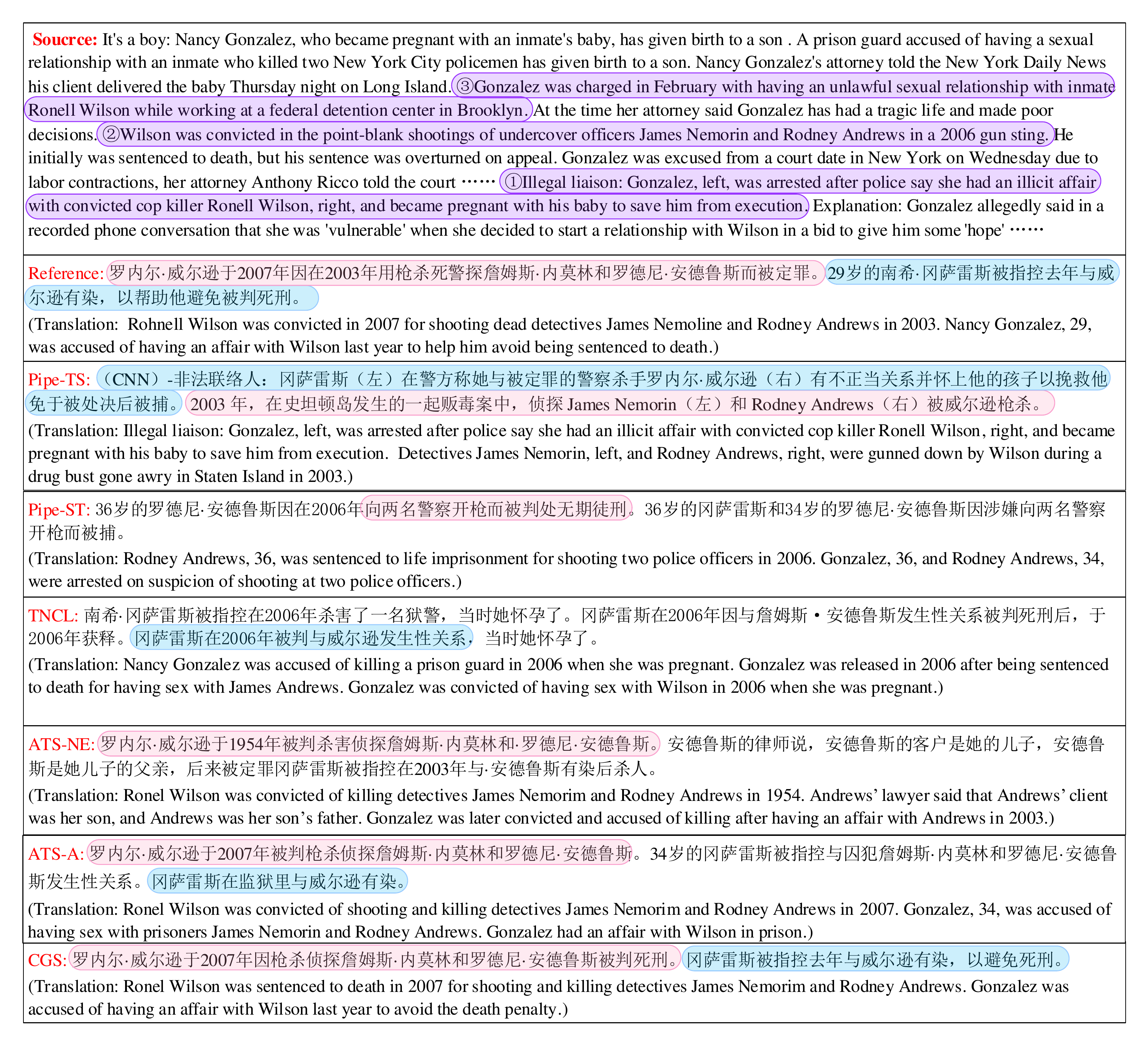}
   \caption{An example of generated En2Zh summaries and its human-corrected reference. The sentences marked with purple boxes are the three most important sentences determined by AGC algorithm. The sentences in summaries marked with the same color box contain similar information. The generated summaries are provided with an English translation for readers to better understand. }
 \label{pic:output}
\end{figure*}

We present a case study of a generated En2Zh summaries in Figure~\ref{pic:output}.
It can be seen that almost all information in the reference comes from the three most important sentences chosen by AGC algorithm and only CGS correctly and concisely summarizes the source article. This reveals the effectiveness of AGC algorithm and its ability to correctly locate important sentences.

We find that the generated summaries of all baselines except for Pipe-TS  more or less capture some information in the reference, but none of them are fully correct and complete.
Since Pipe-TS utilizes Google translator and an extractive summarization method, the fluency and informativeness of its generated summary are obvious better than those of other baselines. But it's summary is less concise than CGSs' and is obvious ``translationese''.

On the other hand, the repeated information like ``Gonzale, pregnant, Andrews, affair, etc.'' in the summaries generated by Pipe-ST, TNCL, ATS-NE and ATS-A imply that the model usually tends to generate repeated information which appears in the reference to improve the recall rate.
 In particular, all generated summaries omit or misrepresent the information of ``Gonzale's age''. Although the summary generated by CGS does not correspond to every word of the reference, it captures all the important points and expresses them in other ways.

In conclusion, our CGS can produce a summary that is more consistent with the reference than baselines.

\section{Discussion and Conclusion}
Cross-lingual automatic summarization is a meaningful and useful research task because it can greatly shorten the time of manual writing and can provide a tool for readers to quickly grasp the gist of an article written in an unfamiliar language. This paper has proposed an end-to-end CLS framework, called ``ClueGraphSum'', to implement the automatic generation of high-quality cross-lingual summaries. The main theoretical implications of this study are as follows:
\begin{itemize}
\item Exploring the leadership role of key clues for summarization in CLS task and proposing a clue-guided cross-lingual summarization method to directly understand and summarize the bilingual semantics.
\item Proposing a novel article construction algorithm (denoted as AGC) that takes sentences as nodes and links the nodes which share common key clues.
\item Proposing a new node weight calculation method to locate important sentences based on AGC algorithm.
\item Exploring the impact of article length to the neural net model and finding that the performance of models decreases with the increase of article length.
\item Exploring the performance of different models on translated datasets and our hand-written CLS dataset.
\end{itemize}

Our proposed method has been evaluated by both automatic evaluation metrics (ROUGE, METEOR, MoverScore) and human judgment (conciseness, informativeness and fluency). Experimental results show that our proposed method can take advantage of key clues to exceed baselines and these clues working with the constructed article graph help to achieve better robustness to the article length than strong baselines.

In terms of practical implications, our research can convert articles in various languages into summaries in the language familiar to readers, providing students, officers and researchers with quick access to the non-native information. This can help them quickly filter and locate the required part in the overwhelming global information, speeding up the process of learning, working and researching.

In the future work, we would like to supplement and reorganize these clues into meta-events to eliminate redundant information and strengthen the universal applicability in minor languages by transferring the knowledge on the cross-lingual pre-training model.

\bibliographystyle{IEEEtranS}
\bibliography{References_backfornew}

\begin{appendices}
\section{CyEn2ZhSum dataset}
 \label{ap:1}
\begin{figure} [!htbp]
  \centering
  \includegraphics[width=1\textwidth]{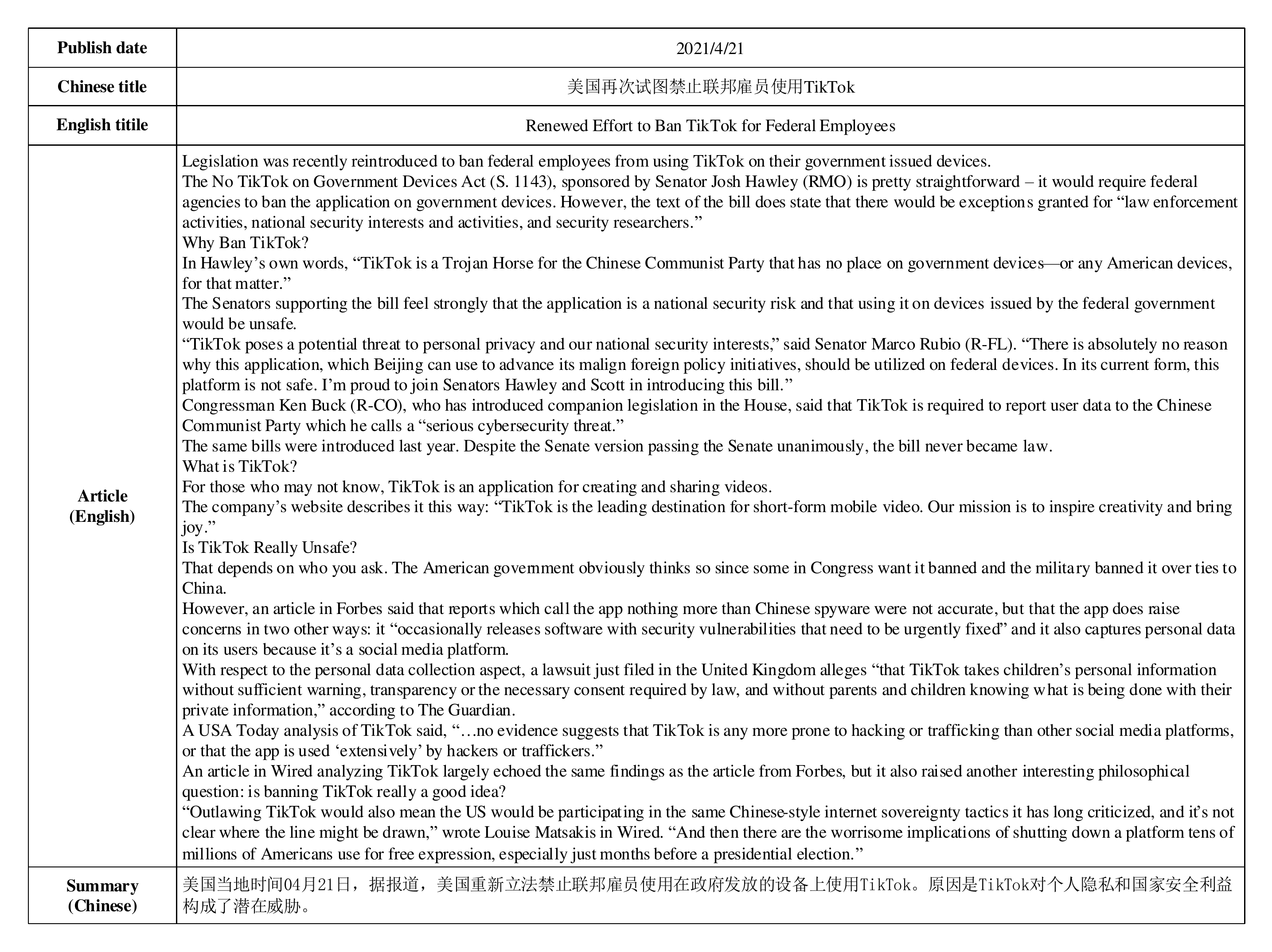}
  \caption{An example of CyEn2ZhSum dataset}
 \label{pic:cyberexample}
\end{figure}

 The example of CyEn2ZhSum dataset is shown in Figure ~\ref{pic:cyberexample}.

\end{appendices}

\end{document}